\documentclass[conference]{IEEEtran}
\pdfoutput=1
\usepackage{cite}
\usepackage{amsmath,amssymb,amsfonts}
\usepackage{algorithmic}
\usepackage{graphicx}
\usepackage{textcomp}
\usepackage{booktabs}
\usepackage{algorithm,algorithmic}

\usepackage{xspace} %
\usepackage{multirow}

\usepackage{adjustbox}  %

\renewcommand{\algorithmicrequire}{\textbf{Input:}}  %
\renewcommand{\algorithmicensure}{\textbf{Output:}} %

\renewcommand{\b}{{\bf b}}

\newcommand{\h}{{\bf h}}
\renewcommand{\k}{{\bf k}}

\newcommand{\w}{{\bf w}}
\newcommand{\x}{{\bf x}}
\newcommand{\y}{{\bf y}}

\newcommand{\I}{{\bf I}}

\newcommand{\K}{{\bf K}}
\renewcommand{\L}{{\bf L}}

\newcommand{\Ocal}{\mathcal{O}}

\newcommand{\Lcal}{\mathcal{L}}

\newcommand{\W}{{\bf W}}

\newcommand{\Mcal}{{\mathcal{M}}}

\newcommand{\bepsilon}{\boldsymbol{\epsilon}}

\newcommand{\btheta}{\boldsymbol{\theta}}

\newcommand{\bxi}{\boldsymbol{\xi}}
\newcommand{\bSigma}{\boldsymbol{\Sigma}}

\newcommand{\bgamma}{\boldsymbol{\gamma}}

\newcommand{\bmu}{\boldsymbol{\mu}}
\newcommand{\1}{{\bf 1}}
\newcommand{\0}{{\bf 0}}

\newcommand{\ben}{\begin{enumerate}}
\newcommand{\een}{\end{enumerate}}

\newcommand{\argmin}{\operatornamewithlimits{argmin}}
\newcommand{\argmax}{\operatornamewithlimits{argmax}}

\newcommand{\EE}{\mathbb{E}}

\newcommand{\cmt}[1]{}

\newcommand{\bXi}{\boldsymbol{\Xi}}

\renewcommand{\x}{\textbf{\textit{x}}}  %
\renewcommand{\w}{\textbf{\textit{w}}}  %
\renewcommand{\y}{\textbf{\textit{y}}}  %
\renewcommand{\k}{\textbf{\textit{k}}}  %
\renewcommand{\b}{\textbf{\textit{b}}}  %

\newcommand{\ours}{BoA-PTA\xspace}

\usepackage{color}

\def\BibTeX{{\rm B\kern-.05em{\sc i\kern-.025em b}\kern-.08em
    T\kern-.1667em\lower.7ex\hbox{E}\kern-.125emX}}
\begin{document}

\title{BoA-PTA, A Bayesian Optimization Accelerated Error-Free SPICE Solver for IC Design\\}
\title{BoA-PTA, Bayesian Optimization Accelerated Error-Free SPICE Simulations \\} 
\title{BoA-PTA, A Bayesian Optimization Accelerated Error-Free SPICE Solver}

\author{
\IEEEauthorblockN{1\textsuperscript{st} Wei W. Xing}
\IEEEauthorblockA{\textit{
		School of Integrated Circuit Science and Engineering} \\
\textit{Beihang University}\\
Beijing, China \\
wxing@buaa.edu.cn}
\and
\IEEEauthorblockN{2\textsuperscript{nd} Xiang Jin}
\IEEEauthorblockA{\textit{
		School of Integrated Circuit Science and Engineering} \\
	\textit{Beihang University}\\
	Beijing, China \\
	jinxiang1114@163.com}
\and
\IEEEauthorblockN{3\textsuperscript{rd} Yi Liu}
\IEEEauthorblockA{\textit{Department of Computer Science and Technology
		} \\
\textit{University of Petroleum-Beijing}\\
Beijing, China \\
enchantedlllll717@163.com}
\and
\IEEEauthorblockN{4\textsuperscript{th} Dan Niu}
\IEEEauthorblockA{\textit{School of Automation} \\
	\textit{Southeast university}\\
	Changsha, China \\
	danniu1@163.com}
\and
\IEEEauthorblockN{5\textsuperscript{th} Weishen Zhao}
\IEEEauthorblockA{\textit{
		School of Integrated Circuit Science and Engineering} \\
	\textit{Beihang University}\\
	Beijing, China \\
	wxing@buaa.edu.cn}
\and
\IEEEauthorblockN{6\textsuperscript{th} Zhou Jin}
\IEEEauthorblockA{\textit{Department of Computer Science and Technology
	} \\
	\textit{University of Petroleum-Beijing}\\
	Beijing, China \\
	jinzhou@cup.edu.cn}
}

\makeatletter
  \def\UTFviii@defined#1{%
    \ifx#1\relax
      !!FIXME!!%
    \else
      \expandafte‌​r#1%
    \fi
  }
\makeatother

\maketitle

\begin{abstract}
One of the greatest challenges in IC design is the repeated executions of computationally expensive SPICE simulations, particularly when highly complex chip testing/verification is involved. Recently, pseudo transient analysis (PTA) has shown to be one of the most promising continuation SPICE solver.
However, the PTA efficiency is highly influenced by the inserted pseudo-parameters.
In this work, we proposed \ours, a Bayesian optimization accelerated PTA that can substantially accelerate simulations and improve convergence performance without introducing extra errors.
Furthermore, our method does not require any pre-computation data or offline training. 
The acceleration framework can either be implemented to speed up ongoing repeated simulations immediately or to improve new simulations of completely different circuits.
\ours is equipped with cutting-edge machine learning techniques, e.g., deep learning, Gaussian process, Bayesian optimization, non-stationary monotonic transformation, and variational inference via reparameterization.
We assess \ours in 43 benchmark circuits against other SOTA SPICE solvers and demonstrate an average 2.3x (maximum 3.5x) speed-up over the original CEPTA.
\end{abstract}


With increasing degrees of the integration of modern integrated circuits (IC), 
the reliability of a chip design is improved via a time-consuming verification process before it can be taped-out～\cite{chen2017parallel}.
The verification mainly verifies whether a designed IC is physically feasible and robust by Monte-Carlo analysis～\cite{akbari2018input}, dynamic timing analysis~\cite{paul2002testing}, and analog circuit synthesis～\cite{zhang2020efficient}, all of which require repeated executions of an expensive SPICE (simulation program with integrated circuit emphasis) simulation (due to the large scale of an IC design)～\cite{negel1975computer}.
This poses a great challenge as the verification can take up to $80\%$ of the development time in an IC design~\cite{wang2009electronic}.

Due to its recent fast development, machine learning and other statistical learning methods have been utilized to resolve this challenge~\cite{huang2021machine}. For instance, Bayesian optimization~\cite{zhang2020efficient}, multi-fidelity modelling~\cite{zhang2019efficient}, and computing budget allocation~\cite{liu2010an} are proposed to accelerate repeated simulations.
Despite being efficient, direct machine learning implementations rely on a large amount of pre-computed data to work. Furthermore, almost all machine learning based methods provide no error bounds in any forms, putting the verification process in great risk.
Thus, the machine learning methods are mainly used in academic research rather than industrial applications.

A ``first principal'' way to reduce the computational expense is to improve the SPICE efficiency.
A SPICE solves nonlinear algebraic equations or differential algebraic equations that are constructed on a circuit base on Kirchhoff's current law (KCL) and Kirchhoff's voltage law (KVL)\cite{gunther1995dae}.
The solution provides direct current (DC) analysis, which supports other detailed analysis, e.g., transient analysis and small signal analysis~\cite{udave2012dc}.
A general SPICE utilizes Newton-Raphson (NR) iteration and some continuation methods, e.g., Gmin stepping~\cite{DBLP:journals/tcad/RoychowdhuryM06} and source stepping~\cite{ter2012robust}, to solve the nonlinear equations due to their fast convergence properties.
However, they may fail to converge when the circuit scale is sufficiently large and especially with strong nonlinearity design, which brings high loop gain, positive feedback, or multiple solutions~\cite{DBLP:journals/tcad/UshidaYNKI02}. 
This challenge is well resolved by pseudo-transient analysis (PTA)~\cite{rezvani2017synthesis}, which inserts constant pseudo capacitors and inductors to original circuits. 
However, PTA can cause oscillation issues.
Damped PTA (DPTA)~\cite{wu2014pta}, exploits a numerical integration method with artificially enlarged damping effect to deal with oscillation;
Ramping PTA (RPTA)~\cite{jin2015effective} ramps up voltage sources instead of inserting the pseudo-inductors to suppress fill-ins. 
Compound element PTA (CEPTA)~\cite{DBLP:journals/ieicet/YuISHH07} has demonstrated a strong capability to eliminate oscillation while maintain a high efficiency. 
As to our knowledge, till now, there has been no literature showing effective (solver) parameter strategies for CEPTA acceleration.

\begin{figure}[t]
    \centering
    \vspace{-0.1in}
    \includegraphics[scale=0.33]{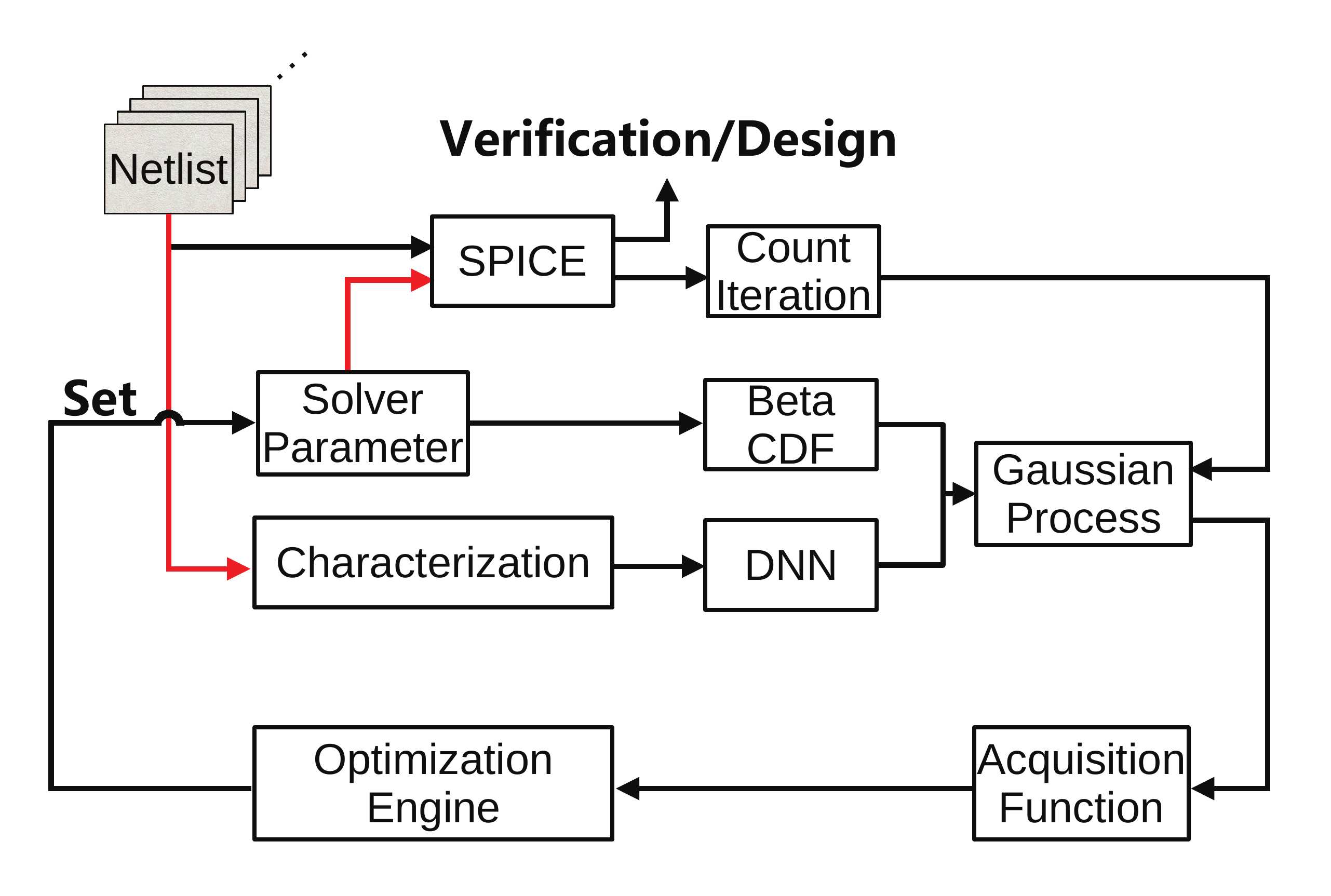}
    \vspace{-0.2in}
    \caption{The proposed \ours\ framework.}
    \label{figOverview}
    \vspace{-0.1in}
\end{figure}

To harness the power of modern machine learning and meanwhile retains accuracy reliability of a SPICE simulation,
we aim to equip the state-of-the-art SPICE solver, CEPTA, with machine learning power.
To this end, 
we propose \ours, a \underline{B}ayesian \underline{o}ptimization error-free \underline{a}cceleration framework using CE\underline{PTA} (Fig.~\ref{figOverview}).
Specifically, we introduce a Bayesian optimization (BO) to select PTA solver parameters as an optimization problem.
To extend the capability for different circuits, we utilize a special netlist characterization and a deep neural network (DNN) for netlist feature extraction.
To further improve \ours for the highly nonlinear optimization problem, we introduce a Bayesian hierarchical warping BetaCDF, which overcomes the stationary limitation of a general BO without complicating the geometry via a monotonic bijection transformation. 
Parameters of the warping BetaCDF are integrated out using variational inference combined with reparameterization to avoid overfitting. 
Lastly, the optimization constraint and scale are handled by a log-sigmoid transformation.
We highlight the novelty of \ours as follows,
\begin{enumerate}
    \item As far as the authors are aware, \ours is the first machine learning enhanced SPICE solver.
    \item \ours provides error-free accelerations and improves convergence performance for SPICE solvers.
    \item \ours requires no pre-computed data. It can accelerate ongoing repeated simulations or to improve new simulations of completely different circuits.
    \item \ours is equipped with cutting-edge machine learning techniques: deep learning for netlist feature extractions, BetaCDF for non-stationary modelling, and variational inference to avoid overfitting.
    \item \ours shows an average 2.2x (maximum 3.5x) speed-up on 43 benchmark circuit simulations and Monte-Carlo simulations.
\end{enumerate}
We implement our acceleration framework for CEPTA due to its urgent need for solver parameter tuning. Nevertheless, our method is ready to combine with other SPICE solver.
As a very first work of machine learning enhanced SPICE, we hope this work can inspire interesting machine learning enhanced EDA tool from different perspectives.

The rest of the paper is organized as follows. 
In Section 2, review the background of PTA and BO.
In Section 3, \ours is derived with motivations and details.
In Section 4, we assess \ours on 43 benchmark circuit simulations for different tasks.
We conclude this work in Section 5.

\section{BACKGROUND AND PRELIMINARIES}
\subsection{SPICE Simulations Via PTA}
\begin{figure}[htbp]
    \centering
    \vspace{-0.15in}
    \includegraphics[scale=0.65]{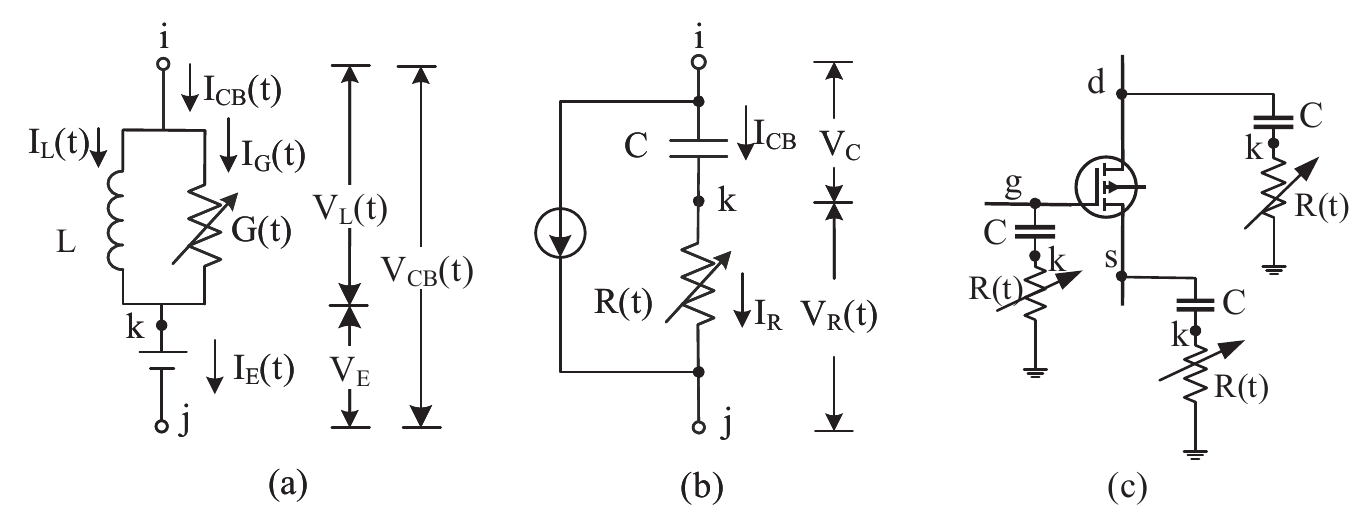}
    \vspace{-0.2in}
    \caption{Inserted pseudo-elements ((a)GVL, (b,c)RVC) and their embedding positions in CEPTA. }
    \label{fig_pta1}
\end{figure}

\begin{figure}[htbp]
    \centering
    \vspace{-0.2in}
    \includegraphics[scale=0.7]{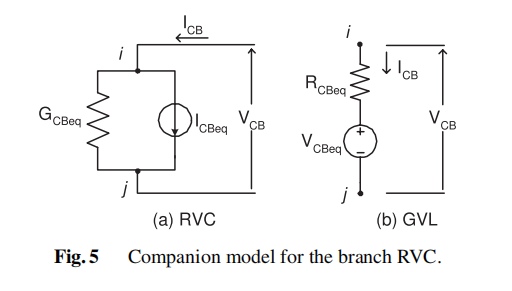}
    \vspace{-0.2in}
    \caption{Equivalent circuit for the branch RVC and GVL.}
    \label{fig_pta2}
\end{figure}

When the commonly used NR and practical continuation methods fail to converge in a SPICE, the PTA is implemented as an alternative because it provides robust solutions to nonlinear algebraic equations from modified nodal analysis (MNA).
PTA works by inserting certain dynamic pseudo-elements into the original circuits. 
As shown in Fig. \ref{fig_pta1}, CEPTA inserts a GVL branch into an independent voltage source in serial (Fig. \ref{fig_pta1}(a)), a RVC branch into an independent current source in parallel (Fig. \ref{fig_pta1}(b)), and a transistors between each node to ground (Fig. \ref{fig_pta1}(c)).
The RVC branch is composed of a constant capacitor $C$ connected in serial with a time-variant resistor $R(t)$ whereas the GVL branch is composed of a constant inductor $L$ connected in parallel with a time-variant conductance $G(t)$. 
With the pseudo-elements inserted, the differential algebraic equations \eqref{e1} is solved with an initial guess $u_0$ until converge
\begin{equation}\label{e1}
    g(u(t),\dot{u}(t),t) =0, \ \
    R(t)=R_0e^{t/\tau}, \ \ G(t)=G_0e^{t/\tau}.
\end{equation}
The converged solution (when $\dot{u}(t)=0$) is the solution to the original circuit. 
The insertion of compound elements brings additional nodes (such as node $k$ in Fig. \ref{fig_pta1}(b)) that require extra computations.
To avoid enlarging the size of the Jacobian matrix induced by additional nodes, CEPTA can be implemented in an equivalent way, where the equivalent circuits are shown as Fig.~\ref{fig_pta2}.
The equivalent equation for branch RVC at time $t^{n+1}$ after discretization by the backward Euler is
\begin{equation}\label{e5}
    I_{CB}^{n+1}=G_{CBeq}V_{CB}^{n+1}+I_{CBeq},
\end{equation}
where $G_{CBeq}^{-1}={h^{n+1}}/{C}+R(t^{n+1})$ and $I_{CBeq}=G_{CBeq}(I_{CB}^nR(t^n)-V_{CB}^n)$.
Similarly, the equivalent equations for branch GVL at time point $t^{n+1}$ can be obtained by
\begin{equation}\label{e8}
  V_{CB}^{n+1}=R_{CBeq}I_{CB}^{n+1}+V_{CBeq},
\end{equation}
where,
$R_{CBeq}^{-1}={h^{n+1}}/{L}+G^{n+1}$ and $V_{CBeq}=R_{CBeq} (-I_{CB}^n+G^n(V_{CB}^n-E))+E$.
Despite CEPTA's great success, its performance is highly influenced by the 
inserted pseudo elements, i.e., the values of inserted pseudo capacitor, inductor, and the initial values of resistor and conductance.
It is thus important to quickly find a set of optimal inserted pseudo elements that accelerates the convergence and thus the repeated SPICE simulations.
The circuit-dependent and sensitive property makes solver parameter tunning an still open challenge.

\subsection{Problem Formulation}
Consider a CEPTA solver $g$ with solver parameters $\x$ (indicating the value of inserted capacitor, inductor, resistor and conductance)
that operates on a netlist file denoted as $\bxi$ and generate the steady state $\textbf{\textit{u}} = g(\x,\bxi)$.
We are interested in reducing the number of iteration, denoted as $\eta(\x,\bxi) + \varepsilon$, for $g(\x,\bxi)$.
Here
$\varepsilon$ captures the model inadequacy and randomness that are not fully captured by $\x$ and $\bxi$.
We aim to seek a function
\begin{equation}
    \x^*(\bxi) = \argmin_{\x \in \mathcal{X}} \eta(\x,\bxi),
\end{equation}
where $\x^*(\bxi)$ is the optimal CEPTA solver parameters for any given netlist $\bxi$ and $\mathcal{X}$ is the feasible domain for $\x$.

\subsection{Bayesian Optimization}
Bayesian optimization (BO) is an optimization framework generally for expensive black-box functions that are noisy or noise-free \cite{mockus2012bayesian}.
Since the black-box function is expensive to evaluate, it is approximated by a probabilistic surrogate model, which provides useful derivative information for the classic optimization approaches.
The surrogate model is a data-driven probabilistic regression that is calibrated to fit the black-box function with available data.
Thus, we need as many data from the black-box function. 
Since the data are expensive to collect, we need to have a good strategy to update the surrogate model to gain maximum improvement with each new observation. This is known as the exploration.
Keep in mind that the ultimate goal is the optimization of the black-box function rather than fitting a surrogate; this is known as the exploitation.
The tradeoff between exploration and exploitation is handled by the acquisition function, which should reflect our reference for the tradeoff.

\subsection{Gaussian Process}
Gaussian process (GP) is a common choice for the surrogate model of BO due to its model capacity for complex black-box function and for uncertainty quantification, which naturally quantifies the tradeoff. We briefly review GP in this section.

For the sake of clarity, let us consider a case where the circuit is fixed and its index $\bxi$ is thus omitted. 
Assume that we have observation $y_i=\eta({\x}_i) + \varepsilon$, $i=1,\dots, N$ and design points ${\x}_i$, where $y$ is the (determined) iteration number needed for convergence.
In a GP model we place a prior distribution over $\eta({\x})$ indexed by ${\x}$:
\begin{equation}\label{scalarem}
\eta({\x})| \pmb{\theta}\sim {\cal GP}\left(m({\x}),k({\x}, {\x}'|\pmb{\theta})\right),
\end{equation}
with mean and covariance functions:
\begin{equation}
\begin{array}{ll}
m_0({\x}) &=\mathbb{E}[\eta({\x})],\\
k({\x},{\x}'|\pmb{\theta}) &=\mathbb{E}[(\eta({\x})-m_0({\x}))(\eta({\x}')-m_0({\x}'))],
\end{array}
\end{equation}
in which $\mathbb{E}[\cdot]$ is the expectation operator. The hyperparameters $\pmb{\theta}$ are estimated during the learning process. The mean function can be assumed to be an identical constant, $m_0({\x})\equiv m_0$, by virtue of centering the data. Alternative choices are possible, e.g.,  a linear function of ${\x}$, but rarely adopted unless {\/\it a-priori} information on the form of the function is available. The covariance function can take many forms, the most common being the automatic relevance determinant (ARD) kernel:
\begin{equation}\label{covfunc}
k({\x}, {\x}'|\pmb{\theta})=\theta_0\exp\left(-({\x}-{\x}')^T\mbox{diag}(\theta_1,\hdots,\theta_l)({\x}-{\x}')\right),
\end{equation}
The hyperparameters $\pmb{\theta}=(\theta_0,\hdots,\theta_l)^T$. $\theta_1^{-1},\hdots,\theta_l^{-1}$ in this case  are called the square correlation lengths. For any fixed ${\x}$, $\eta({\x})$ is a random variable. A collection of values $\eta({\x}_i)$, $i=1,\hdots,N$, on the other hand, is a partial realization of the GP. Realizations of the GP are deterministic functions of ${\x}$. The main property of GPs is that the joint distribution of $\eta({\x}_i)$, $i=1,\hdots,N$, is multivariate Gaussian.
Assuming the model inadequacy $\varepsilon \sim \mathcal{N}(0,\sigma^2)$ is also a Gaussian, with the prior \eqref{scalarem} and available data $\y=(y_1,\hdots,y_N)^T$, we can derivative the model likelihood
\begin{equation} \label{MLE}
    \begin{aligned}
        \Lcal & \triangleq  p(\y|\x,\btheta) = \int_\eta (\eta(\x) + \varepsilon) d \eta
        = \mathcal{N}(\y|m_0 \1, \textbf{K}(\pmb{\theta}) +\sigma^2\I) \\
        &= -\frac{1}{2} \left( {\y} - {m_0} {\bf 1} \right)^T  (\textbf{K}(\pmb{\theta}) +\sigma^2\I) ^{-1} \left( {\y}-{m_0} {\bf 1} \right) \\
        & \quad -\frac{1}{2}\ln|\textbf{K}(\pmb{\theta}) + \sigma^2\I|  - \frac{N}{2} \log(2\pi),
    \end{aligned}
\end{equation}
where the covariance matrix $\K({\btheta})=[K_{ij}]$, in which $K_{ij}=k({\x}_i, {\x}_j|\pmb{\theta})$, $i,j=1,\hdots,N$.
The hyperparameters $\pmb{\theta}$ are normally obtained from point estimates~\cite{kennedy2001bayesian} by maximum likelihood estimate (MLE) of \eqref{MLE} w.r.t. $\btheta$.
The joint distribution of $\y$ and $\eta(\x)$ also form a joint Gaussian distribution with mean value $m_0 \1$ and covariance matrix
\begin{equation}
    \begin{array}{c}\label{cm3}
    \displaystyle \textbf{K}'(\pmb{\theta})=
    \left[ 
        \begin{array}{c|c} \textbf{K}(\pmb{\theta}) + \sigma^2\I & {\k}({\x})  \\ 
        \hline
        {\k}^T({\x})& k({\x}, {\x}|\pmb{\theta})  + \sigma^2
    \end{array}
    \right],
    \end{array}
\end{equation}
in which ${\k}({\x})=(k({\x}_1, {\x}|\pmb{\theta}),\hdots, k({\x}_N, {\x}|\pmb{\theta}))^T$.
Conditioning on  ${\y}$ provides the conditional predictive distribution  at ${\x}$~\cite{rasmussen2006gaussian}:
\begin{equation}\label{postpredA}
\begin{array}{c}
\hat{\eta}({\x})|{\y},\pmb{\theta}\sim \mathcal{N}\left(\mu ({\x}|\pmb{\theta}), v ({\x},{\x}'|\pmb{\theta})\right),
\vspace{2mm}\\
\mu ({\x}|\pmb{\theta})= m_0 {\bf 1} + {\k}({\x})^T \left( \textbf{K}(\pmb{\theta}) + \sigma^2 \I \right)^{-1} \left( \y - {m_0} {\bf 1} \right),
\vspace{2mm}\\
v ({\x}|\pmb{\theta})=  \sigma^2 + k({\x}, {\x}|\pmb{\theta}) -{\k}^T({\x}) \left( \textbf{K}(\pmb{\theta}) + \sigma^2 \I \right)^{-1} {\k}({\x}).
\end{array}
\end{equation}
The expected value $\mathbb{E}[\eta({\x})]$ is given by $\mu({\x}|\pmb{\theta})$ and the predictive variance by $v({\x}|\pmb{\theta})$.

\subsection{Acquisition Function}
For simplicity, let us consider the maximization of the black-box function without particular constraints.
Based on the GP model posterior in \eqref{postpredA},
we can simply calculate the improvements for a new input $\x$ as  $I(\x) = \max( \hat{\eta}(\x) - y^\dagger , 0)$, where $y^\dagger$ is the current optimal and $\hat{\eta}(\x)$ is the predictive posterior in \eqref{postpredA}.
The expected improvement (EI) \cite{jones1998efficient} over the probabilistic space is
\begin{equation}\label{EIfunc}    
    \begin{aligned}
        & EI(\x) = \EE_{\hat{\eta}(\x) \sim \mathcal{N}(\mu(\x), v(\x)}) [\max( \hat{\eta}(\x) - y^\dagger , 0) ] \\
        &= (\mu(\x) - y^\dagger) \psi \left(\frac{ \mu(\x) - y^\dagger}{v(\x)}\right)  
         + v(\x) \phi \left(\frac{\mu(\x) - y^\dagger}{v(\x)} \right),
    \end{aligned}
\end{equation}
where $\psi(\cdot)$ and $\phi(\cdot)$ are the probabilistic density function (PDF) and cumulative density function (CDF) of a standard normal distribution, respectively.
It is clear that the EI acquisition function \eqref{EIfunc} favors regions with larger uncertainty or regions with larger predictive mean values and naturally handles the tradeoff between exploitation and exploration.
The candidates for next iteration is selected by
\begin{equation}
    \argmax_{\x \in \mathcal{X}} EI(\x),
\end{equation}
which is normally optimized with classic non-convex optimizations, e.g., L-BFGS-B.

Rather than looking into the expected improvement, we can approach the optimal by exploring the areas with higher uncertainty towards to the maximum
\begin{equation}\label{eqUCB}
    \argmax_{\x \in \mathcal{X}} \left(\mu(\x) + \beta^{\frac{1}{2}} v(\x) \right),
\end{equation}
where $\beta$ reflects our preference of the tradeoff of exploration and exploitation. This is known as the upper confidence bound (UCB) \cite{srinivas2010gaussian}, which is simple and easy to implement yet powerful and effective. However, the choice for $\beta$ is nontrivial, which hinders its further applications. 

Both EI and UCB acquisition functions try to extract the best from the current status. The max-value entropy search (MES) acquisition function is introduced by \cite{wang2017max} to take a further step to inquire at a location that produce maximum information gain (based on information theory) about the black-box function optimal,
\begin{equation} \label{eq_MES}
    \begin{aligned}
        MES(\x) &= -\EE_{y(\x)}[ h (y^* | \mathcal{D} \cup \{\x, \hat{\eta}(\x)\})] - H(y^*|\mathcal{D})\\
        &= - \EE_{\hat{\eta}^*}[ H(\hat{\eta}(\x)|y^*)] + H(\hat{\eta}(\x)),
    \end{aligned}
\end{equation}
where $y^*$ indicates the black-box function optimal, $\mathcal{D}$ means the current data set, and $H(\hat{\eta}) = -\int p(\hat{\eta}) \log(p(\hat{\eta})) d \hat{\eta} $ is the entropy for $p(\hat{\eta})$. The first term in \eqref{eq_MES} is generally achieved using sampling method whereas the second one has a closed-form solution. The readers are referred to \cite{wang2017max} for more details. 

BO is an active research area and there are many other acquisition functions, e.g., knowledge gradient \cite{scott2011the} and predictive entropy search \cite{hernandez-lobato2014predictive}. Ensembles of multiple acquisition function are also possible \cite{lyu2018batch}.
In this work, we focus on BO accelerated SPICE and we test it with EI, UCB, and MES. However, our method can be combined with any existing acquisition strategy.

\section{Proposed Boe-PTA}

\subsection{Circuits Characterization Via Deep Learning}    \label{sec_dnn}
The most challenging part in this work is the characterization of the circuit  where the SPICE solver is executed on. 
Recently, machine learning techniques have been implemented in the EDA community to accelerate the design/verification process \cite{huang2021machine}. 
Directly introducing a powerful model such as deep learning that directly uses netlist as inputs is feasible in some cases \cite{ma2019high}. 
However, this approach is unlikely to address our problem because we do not have a large amount of data nor great computational budge for model training. 
Even if we have, the overwhelming computational overhead required will make the approach impractical for real problems. 
Instead, we follow \cite{zhang2019bayesian} and use the
seven key factors (the number of nodes, MNA equations, capacitors, resistors, voltage sources, bipolar junction transistor, and MOS field-effect transistor) to characterize a netlist as raw inputs for \ours.
These features are denoted as a column input $\bxi$.
A GP with commonly used kernel (e.g., \eqref{covfunc}) is unlikely to be able to capture the complex correlations between different netlists.
The DNN has shown to be a powerful automatic feature extraction for various practical applications \cite{lecun2015deep}. 
Thus, we further introduce a deep learning transform as an automatic feature extraction for $\bxi$, i.e., $\Phi(\bxi)$, before the GP surrogate.
\begin{equation}
    \Phi(\bxi) = \h( \W^{l} \Phi^{l}(\bxi)  + \b^l),
\end{equation}
where $\Phi^{l}(\bxi) = \h( \W^{l-1} \Phi^{l-1}(\bxi)  + \b^{l-1})$, $\Phi^{1}(\bxi)=\bxi$, and $\h(\cdot)$ is an element-wise nonlinear transformation known as the active function in this scenario. 
This is a classic DNN structure known as the multiple layer perception (MLP), which is commonly used to process features in a deep model.
In this work, we use the same dimension of $\bxi$ to be the output dimension of $\Phi(\bxi)$. The extracted features are then passed to a GP for further feature selections by 
an ARD kernel and for model predictions. 
The DNN with the follow-up kernel together can be seen as a kernel that learning complex correlation automatically through DNN. For this reason, this approach is also known as deep kernel learning \cite{wilson2016deep}.

\subsection{Non-Stationary Gaussian Process For CEPTA}
The efficiency and effectiveness of BO is highly determined by the accuracy of the surrogate model;
for a GP model, its model capacity is largely influenced by the choice of the kernel function.
Consider the function $\eta(\x, \bxi)$ for a fixed $\bxi$, according to our experiments, $\eta(\x, \bxi)$ is a highly nonlinear function w.r.t. $\x$, making the commonly used stationary ARD kernel ineffective for modeling such a complex function.

Unlike the previous section where the complex correlations of $\bxi$ can be captured automatically using a complex model such as DNN~\cite{zhang2019bayesian}, latent space mapping \cite{bornn2012modeling}, GP \cite{adams2008gaussian},
modeling of $\x$ requires extra cares because:
1) despite the strong model capacity, introducing a complex model is likely to introduce extra model parameters (particularly when a DNN is implemented), which makes the model training difficult and potentially requires more data for the surrogate to perform well.
2) Even worse, introducing another complex model can complicate the geometry, making the optimization of the non-convex acquisition function w.r.t. $\x$ more difficult.
Note that the DNN we implement in Section \ref{sec_dnn} does not suffer from this issue because it is not involved in the optimization of acquisition function. We will show the details in later sections.

For modelling $\x$,
we believe the rule of thumb is to follow the Occam's razor and introduce a simple yet effective transformation for the solver parameter $\x$. To this end, we follow the work of \cite{slaets2013warping} and introduce a bijection Beta cumulative density function (BetaCDF),
\begin{equation} \label{eqBetaCDF}
    w_d(x_d) = \int_0^{x_d} \frac{u^{\alpha_d-1} (1-u)^{\beta_d-1} }{B(\alpha_d,\beta_d)} d u,
\end{equation}
where $\alpha_d$ and $\beta_d$ is the positive functional parameters and $B(\alpha_d,\beta_d)$ is the normalization constant. This transformation is monotonic (thus does not complicate the optimization geometry) and it comes with only two extra parameters for each input dimension.
To further reduce the probability of overfitting with $w_d(x_d)$, we  use a hierarchical Bayesian model by placing priors
\begin{equation}
    \log(\alpha_d) \sim \mathcal{N} (\mu_a^d, \sigma_a^d), \quad  \log(\beta_d) \sim \mathcal{N} (\mu_b^d, \sigma_b^d), 
\end{equation}
for the BetaCDF.
The introduced hyperparameters $\{\alpha_d, \beta_d\}_{d=1}^D$ can be obtained via point estimations.
To avoid overfitting, \cite{slaets2013warping} integrate them out by using Markov chain Monte Carlo slice sampling, which significantly increases the model training time and will make the acceleration via \ours impractical because the BO itself consumes too much computational resource.

In this work, reparameterization trick \cite{kingma2014auto} is utilized to conduct a fast posterior inference for $\{\alpha_d, \beta_d\}_{d=1}^D$, which is later integrated out.
We use a log-Gaussian variational posterior $ \log( \bgamma ) \sim \mathcal{N}(\bmu_\gamma, \bSigma_\gamma)$,
where $\bgamma = [\alpha_1,\dots,\alpha_D, \beta_1, \cdots, \beta_D]^T $. This formulation allows us to capture the complex correlation between any $\alpha_d$ and $\beta_{d'}$.

\subsection{Handling Constraints And Scales}
The CEPTA solver parameters are practically in the range of $[10^{-7}, 10^{7} ]$. This poses two challenges. First, it turns the unconstrained optimization into a constrained one that requires extra cares. Second, in its original space, $[10^{-7}, 0 ]$ takes almost zero volume of the whole domain $[10^{-7}, 10^{7} ]$. This makes an optimization either ignore the $[10^{-7}, 0 ]$ range completely or fail to search the whole domain with small searching step.

To resolve these issues simultaneously, we introduce a log-sigmoid transformation,
\begin{equation}
    x_d = \left( 7 \cdot \text{sigmoid} (z_d) \right)^{10},
\end{equation}
where $\text{sigmoid} (z_d) = {1}/({1+\exp(z_d)})$ is the sigmoid function. In this equation, the base-10 logarithm scales $x_d$ such that the optimization focuses on the magnitude of $x_d$ rather the particular value whereas the sigmoid function naturally bounds $x_d$ to the range of $[10^{-7}, 10^{7}]$.
This log-sigmoid transformation is applied to each $x_d$ independently. When the optimization of acquisition is conducted, it is optimized w.r.t. $z_d$ instead of $x_d$. 
Note that this log-sigmoid transformation does not change the monotone of $\eta(\x,\bxi)$ nor affect the non-stationary transformation \eqref{eqBetaCDF}, which is designed to tweak the space $\mathcal{X}$ to resolve the non-stationary issue.

\subsection{Boa-PTA Training and Updating}
Given observation set $\{\x_i,\bxi_i,y_i\}_{i=1}^N$,
the hyperparameters $\btheta$, the DNN $\Phi$ parameters $\{\W^l,\b^l\}_{l=1}^L$, and the variational posterior parameters $\{\bmu_\gamma, \bSigma_\gamma\} $ are updated using gradient descent.
The joint model likelihood $\Lcal$ is the same as \eqref{MLE} but with a composite covariance kernel function 
\begin{equation} \label{eq_kernel2}
    k\left( \left[\w(\x), \Phi(\bxi )\right], \left[\w(\x'), \Phi(\bxi')\right] \right).
\end{equation}
The GP hyperparameters are updated by maximizing $\Lcal$ as in a general GP.
The DNN $\Phi$ parameters $\W^l$ and $\b^l$ are updated by
\begin{equation}
    \frac{\partial \Lcal}{ \partial  \W^l} = \frac{\partial \Lcal}{ \partial k} \cdot \frac{\partial k}{ \partial \Phi} \cdot \frac{\partial \Phi}{ \partial \W^l },
    \quad 
    \frac{\partial \Lcal}{ \partial \b^l} = \frac{\partial \Lcal}{ \partial k} \cdot \frac{\partial k}{ \partial \Phi} \cdot \frac{\partial \Phi}{ \partial \b^l },
\end{equation}
where $\frac{\partial k}{ \partial \Phi}$ represents the gradient of the model log-likelihood w.r.t. the DNN.
For the variational posterior $\bgamma$, we use the parameterization trick to update the variational parameters $\bmu_\gamma$ and $\bSigma_\gamma$.
Specifically,
we sample $\hat{\bgamma}_i$ for $i=1,\cdots,S$ by 
\begin{equation}
    \hat{\bgamma}_i = \exp \left( \bmu_\gamma + \bepsilon \cdot \L \right),
\end{equation}
where $\bepsilon \sim \mathcal{N}(\0, \I)$ is sampled from i.i.d. standard normal distributions and $\L \L^T = \bSigma_\gamma$ is the Cholesky decomposition of $\bSigma_\gamma$.
Given solver parameter $\x$, the output of the BetaCDF warping becomes a distribution. We simplify this process by taking its expectation as the output, we have
\begin{equation}
    \w(\x | \bgamma) \approx \frac{1}{S} \sum_{i=1}^S \w(\x | \bgamma_i).
\end{equation} 
The variational parameters $\bmu_\gamma$ and $\bSigma_\gamma$ can be now updated using back propagation. Taking $\bmu_\gamma$ for instance, we have
\begin{equation}
    \frac{\partial \Lcal}{ \partial \bmu_\gamma} = \frac{1}{S} \sum_{i=1}^S \frac{\partial \Lcal}{ \partial k} \cdot \frac{\partial k}{ \partial \w } \cdot \frac{\partial \w }{ \partial \hat{\bgamma}_i } \cdot \frac{\partial \hat{\bgamma}_i  }{ \partial \bmu_\gamma}.
\end{equation}
In practice, since the surrogate model is updated consequently, we set $S=1$ to save computational resource as in \cite{kingma2014auto}. 
Also, when updating the posterior, we also include the KL distance $\text{KL}(q(\bgamma) || p(\bgamma))$ in the likelihood function.

\subsection{Simplified Optimization Of Acquisition Functions}
Unlike a general BO process where all input parameters of the surrogate model are optimized simultaneously, in our application, we always optimize the solver parameters $\x$ conditioned on a given netlist $\bxi$.
This makes the optimization much easier and faster without repeated forward and backward propagation through the DNN.
Specifically, conditioned on a netlist $\bxi$,
the kernel \eqref{eq_kernel2} can be decomposed as 
$k_1(\w(\x),\w(\x')) \cdot k_2(\Phi(\bxi),\Phi(\bxi'))$, where $k_1$ is an ARD kernel for $\w(\x)$ and $k_2$ for $\Phi(\bxi)$ with their original hyperparameters, due to the separate structure of an ARD kernel.
This decomposition significantly simplifies the optimization of acquisition function because $k_2(\Phi(\bxi),\Phi(\bxi')) $ need to be computed only once until a new target netlist $\bxi_*$ is given.

\subsection{\ours For Solver Parameter Optimization}
Most surrogate model based acceleration techniques require pre-computed data for pseudo-random inputs~\cite{xing2016manifold}.
In contrast, 
\ours can be immediately deployed to explore the potential improvements for a netlist set $\bXi = [\bxi_1,\dots,\bxi_L]$. We call this ``cold start'' because the surrogate has no prior knowledge. 
For this situation, we use a batch iteration scheme to run \ours which is described in Algorithm \ref{algo1}. 

 \begin{algorithm}
 \caption{\ours \ Cold Start}
 \begin{algorithmic}[1]  \label{algo1}
 \renewcommand{\algorithmicrequire}{\textbf{Input:}}
 \renewcommand{\algorithmicensure}{\textbf{Output:}}
 \REQUIRE Netlists $ [\bxi_1,\dots,\bxi_L] = \bXi$, number of epoch $N_{epoch}$
\STATE Execute CEPTA with default setting on any netlist $\bxi_i$
    \FOR {$j = 1$ to $N_{epoch}$}
        \FOR {$i = 1$ to $M$}
        \STATE Update surrogate model $\Mcal$ by maximizing \eqref{MLE} %
        \STATE Update and optimize acquisition function \eqref{EIfunc}, \eqref{eqUCB}, or \eqref{eq_MES} given $\bxi_i$ and get candidate $\x$
        \STATE Execute CEPTA and collect iteration $\eta(\x, \bxi_i)$
        \ENDFOR
    \ENDFOR
    \RETURN Best record of \{$\x^*$, $\eta(\x^*, \bxi_i)$\} for $i=1,\dots,M$; surrogate model $\Mcal$
 \end{algorithmic} 
 \end{algorithm}
It might seem unnecessary to run Algorithm \ref{algo1} because it requires repeated executions of the CEPTA solver and consumes extra computational resources.
Indeed, this process provides little value for the task of solving netlists $\bXi$ for once.
However, \ours provides three significant extended values.
First, it explores the potential improvements for the considered netlists. Unlike a random or grid search scheme, it provides a systematic and efficient way to continuously optimize CEPTA for future usage. 
In practice, those optimal solver parameters can be reused when slight modifications are made to the original netlist, which is a common situation in circuit optimization or yield estimation. We will discuss this further for practical acceleration in Section \ref{secMcAcce}.
Second, for some netlists, CEPTA does not converge with the default solver parameters. In this case, \ours has the potential to seek solver parameters that leads to convergence. This is practically useful for performance improvements for CEPTA.
Third, the process is an effective and efficient way to conduct offline training for the surrogate model to directly predict optimal solver parameters for an unseen circuit/netlist in the future usage. 

\subsection{\ours For Monte-Carlo Acceleration} \label{secMcAcce}
One of the most common situations for repeated SPICE simulation is the Monte-Carlo analysis, where a large number of modest variations of a given netlist are simulated.   
Denote $\mathcal{Q}$ as a netlist sampler which generates a netlist $\bxi$ based on a pre-defined distribution.
Here we propose a possible method for \ours to accelerate such a Monte-Carlo analysis in Algorithm \ref{algo2}.
In this algorithm, we set $y=9999$ as the penalty for a non-convergence case, $2y^*$ as the threshold to halt a solver, $20$ epoch as a convergence threshold. This hyperparameters needs to be adjusted for different situations and computational allowance.
Note that the pre-trained model of Algorithm~\ref{algo1} can be used for Algorithm~\ref{algo2} as a ``warm start'' that provide prior knowledges.

\begin{algorithm}
    \caption{\ours Monte-Carlo Acceleration}
    \begin{algorithmic}[1]  \label{algo2}
    \REQUIRE Netlists sampler $\mathcal{Q}$, number of samples $N_{mc}$
    \STATE Sample a netlist $\bxi$ from $\mathcal{Q}$ and execute $\eta(\x, \mathcal{Q})$
    \STATE Update record of best $\x^*$ and best iteration $y^*$
    \FOR {$i = 1$ to $N_{mc}$}
    \STATE Update surrogate model $\Mcal$ by maximizing \eqref{MLE}
    \STATE Sample a netlist $\bxi$ from $\mathcal{Q}$
    \STATE Optimize acquisition function and get optimal $\x$
    \STATE Execute $\eta(\x, \bxi)$: if SPICE iteration reaches $2 y^*$, stop the execution and set $y=9999$; re-execute $\eta(\x^*, \mathcal{Q})$ and collect results.
    \STATE Update record of best parameters $\x^*$ and iteration $y^*$
    \STATE If $y^*$ has no improvements over 20 iterations, stop BO and use constant $\x^*$ for the rest of simulations.
    \ENDFOR
    \RETURN Monte-Carlo analysis for $\mathcal{Q}$
    \end{algorithmic} 
\end{algorithm}

\subsection{Error Analysis And Computation Complexity}
Unlike many verification/design acceleration solutions that are purely based on machine learning techniques \cite{huang2021machine} introducing unquantified error and uncertainty,
 \ours introduces no extra error or uncertainty.
Specifically, as long as the CEPTA converges, the error is bounded by the error of PTA, which is $(\dot{u}<10^{-12})$ by default. 
\ours thus an error-free approach.
When \ours fails to improve CEPTA and lead to non-convergence situation, we are fully aware of such an error and can roll back to use the default setting.

Once the GP is trained, it only takes $\Ocal(N)$ and $\Ocal(N^2)$ ($N$ is number of observations) for the computation of $\mu(\x)$ and $v(\x)$, respectively.
The complexity of the DNN (depending on the network structure) is approximately $\Ocal(\sum_{l=1}^L M_l^2 )$, where $M_l$ is the number of unite in the hidden layer $l$. The BetaCDF transformation computational cost is negligible.

For the training of a GP, the major computational cost is the matrix inversion $(\K+\sigma^2\I)^{-1} $, which is $\Ocal(N^3)$, and the DNN forward computation for all observations, which is $\Ocal(N\sum_{l=1}^L M_l^2)$.
We can see that \ours scales poorly with $N$, which hinder its further applications.  
In such a case, a variational sparse GP \cite{titsias2009variational} can be implemented to resolve this issue, which is outside the scope of this paper; we thus leave it as a future work.

For practical SPICE simulations that can take up to several hours, \ours brings almost zero computational overhead until the samples grows very large. As discussed above, a sparse GP is then required.

\section{Experimental Results}
\subsection{Benchmark Circuits And Experimental Setups}
Most SPICE solvers have certain advantages on some particular circuits.
To assess \ours thoughtfully, we test it on the circuit simulator benchmark set known as CircuitSim93 \cite{barby1993circuitsim93}, which contains 43 classic circuits including BJT-, MOS2-, and MOS3-type circuits.
We compare \ours to Ngspice, the widely used open source SPICE based on Gmin stepping, and the other PTA family SPICE including PPTA, DPTA, RPTA, and CEPTA.

For \ours, we utilize a DNN of two hidden layers, each of which contains 16 hidden units and a sigmoid activation function.
We use L-BFGS-B with five iterations for both the GP fitting optimization and the acquisition optimization. 
We evaluate \ours with three acquisition functions, i.e., EI, UCB, and MES.
For the UCB, we set the common $\beta=0.1$. The implementation of \ours is based on PyTorch and BoTorch\footnote{https://pytorch.org; \quad  https://botorch.org}.

\subsection{\ours  Acceleration Efficiency} \label{exp1}
\begin{figure}[t]
    \centering
    \includegraphics[scale=0.45]{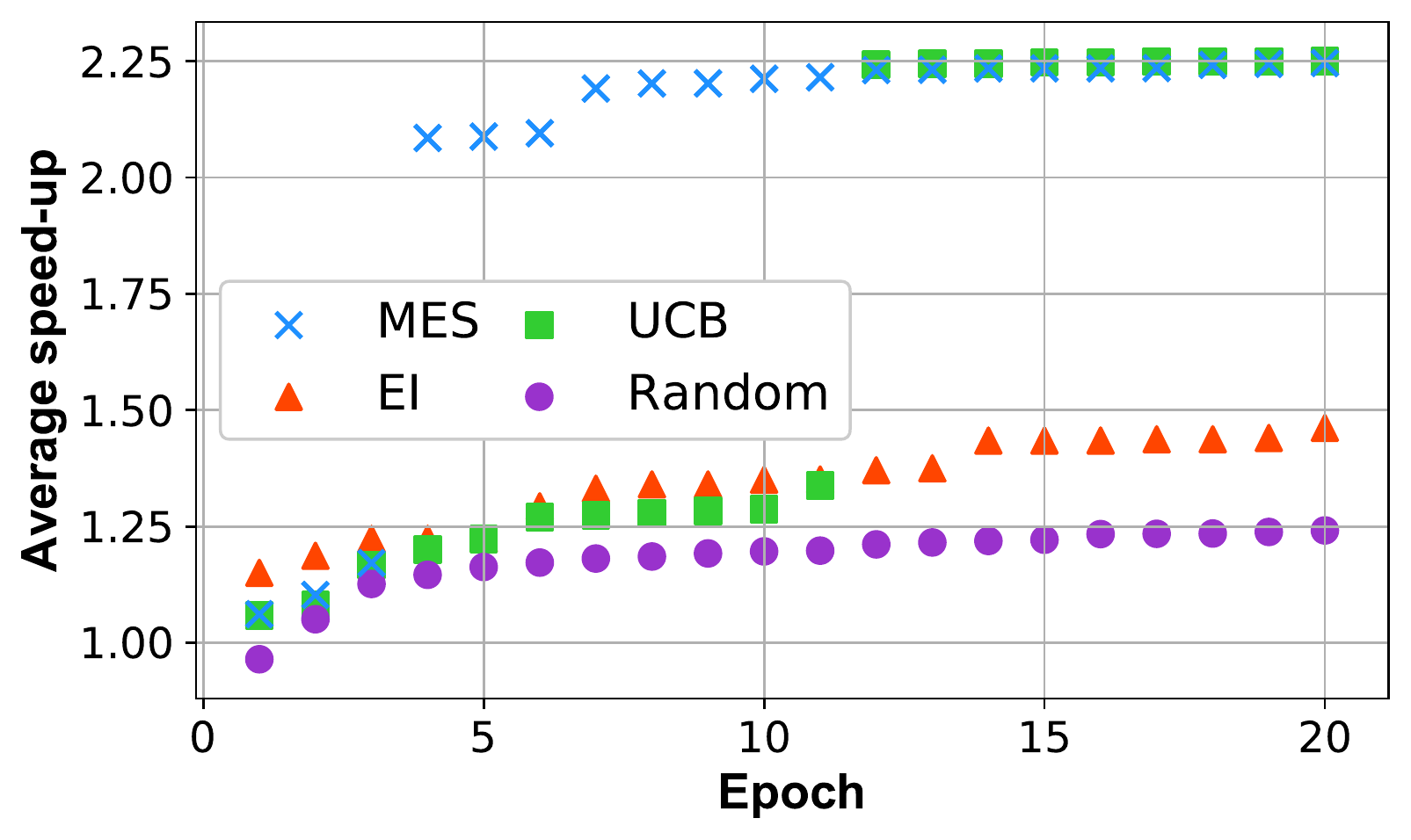}
    \vspace{-0.1in}
    \caption{Average speed-up over 43 simulations for \ours \ with different acquisition function (MES, EI, and UCB) and a random search scheme.}
    \label{figRandom}
    \vspace{-0.1in}
\end{figure}

To show that \ours actually improves the SPICE efficiency, we firstly compare \ours with a vanilla random search method, which run each simulation with randomly sampled solver parameters from a uniform distribution over $\mathcal{X}$. We use Algorithm \ref{algo1} to run \ours for all benchmark simulations with 20 epochs.
This experiment is repeated five times with different random seeds to ensure robustness and fairness of the results.  
After excluding the non-convergence simulations, the average best records of speed-ups over the 43 benchmark simulations (and five repeat tests) are shown in Fig.~\ref{figRandom}.
We can see that \ours with MES clearly outperforms \ours with other acquisition functions with a large margin, particularly with only a few epochs.
The UCB converges to the same speed-up with slightly more iterations. In contrast, the EI struggles to improve quickly. Nevertheless, \ours with any acquisition is clearly better than the random search scheme.

We show the detailed acceleration with 20 epochs for all benchmark simulations
in Table~\ref{tab_iter}.
All DPTA and RPTA results are worse than CEPTA and are not presented due to the limited space.
First thing we notice in Table~\ref{tab_iter} is that \ours outperforms the best PTA method CEPTA in all cases except for pump and reg0, in which the performance are equaled.
This clearly demonstrates that \ours improves CEPTA solver without degeneration.
Another interesting thing to point out is that for the CEPTA non-convergence cases of \{opampal, optrans, ring, gm19\}, \ours makes them converge! This is a particularly useful for PTA based SPICE as they sometime suffer from non-convergence issues.
In general, tunning a PTA solver to converge is extremely difficult because no gradient or space geometry information can be inferred from non-convergence data. 

One may notice that Ngspice also demonstrates a good performance as it outperforms \ours a few times.
This is not surprising because Ngspice utilizes the Gmin stepping, which is particularly good for small-scale simple circuits but scales poorly to large-scale real-world circuits due to non-convergence issues~\cite{780151}. 
We can already see this in Table~\ref{tab_iter} as there are eight non-convergence cases for the Gmin stepping and only four cases for \ours-MES.
In contrast, PTA based methods are known to be robust to large-scale problems but often suffer from slow convergence issues, leading to a large number of iterations.
We highlight that \ours improves CEPTA so much that it can match Gmin in many small-scale benchmark circuits, e.g., ring11 and ab-ac. 
For this reason, we also highlight the best of our results in Table~\ref{tab_iter} to compare among PTA-based methods.

\begin{table}[h]
\caption{Solver iterations for benchmark circuits}
\vspace{-0.1in}
\label{tab_iter}
\begin{adjustbox}{width=0.98\columnwidth,center}
\begin{tabular}{lcccccc}
\toprule
\textbf{bjt} & \textbf{Ngspice} & \textbf{PPTA} & \textbf{CEPTA} & \textbf{MES} & \textbf{EI} & \textbf{UCB} \\
\midrule
astabl & 3359  & 108   & 55    & \textbf{46} & 50    & \textbf{46} \\
bias  & \textbf{86} & N/A   & 839   & \textbf{239} & 294   & 355 \\
bjtff & 116   & N/A   & 169   & 102   & 90    & \textbf{86} \\
bjtinv &  N/A      & 125   & 186   & 90    & \textbf{52} & 53 \\
latch & \textbf{46} & 148   & 130   & 86    & 84    & \textbf{83} \\
loc   &   N/A     & N/A   & N/A   & N/A   & N/A   & N/A \\
nagle & \textbf{117} & 2440  & \textbf{306} & \textbf{306} & \textbf{306} & \textbf{306} \\
opampal & \textbf{168} & 2335  & N/A   & 794   & 866   & \textbf{635} \\
optrans & N/A   & N/A   & 2206  & \textbf{1561} & 2118  & 2283 \\
rca   & \textbf{47} & 76    & 82    & \textbf{55} & 57    & 64 \\
ring11 & \textbf{41} & 102   & 63    & 63    & 63    & \textbf{51} \\
schmitecl & N/A   & 48    & 52    & 45    & 47    & \textbf{44} \\
vreg  & 37    & N/A   & \textbf{22} & \textbf{22} & \textbf{22} & \textbf{22} \\
\midrule
\textbf{mos2} & \textbf{Ngspice} & \textbf{PPTA} & \textbf{CEPTA} & \textbf{MES} & \textbf{EI} & \textbf{UCB} \\
\midrule
ab\_ac & \textbf{53} & N/A   & 90    & \textbf{79} & \textbf{79} & 82 \\
ab\_integ & \textbf{64} & N/A   & 499   & 460   & \textbf{454} & 471 \\
ab\_opamp & 583   & N/A   & 150   & \textbf{121} & 126   & 124 \\
cram  & \textbf{40} & N/A   & 91    & 90    & 91    & \textbf{87} \\
e1480 & 975   & 3213  & 179   & 165   & 134   & \textbf{118} \\
g1310 & 1254  & N/A   & 76    & \textbf{48} & \textbf{48} & 51 \\
gm6   & N/A   & N/A   & 63    & \textbf{42} & 43    & 45 \\
hussamp & \textbf{46} & N/A   & 91    & 88    & 85    & \textbf{82} \\
mosrect & \textbf{44} & 251   & 65    & 54    & \textbf{51} & 53 \\
mux8  & \textbf{25} & 8579  & 122   & \textbf{90} & 93    & 91 \\
nand  & \textbf{25} & N/A   & 83    & 55    & 56    & \textbf{54} \\
pump  & 47    & N/A   & \textbf{22} & \textbf{22} & \textbf{22} & \textbf{22} \\
reg0  & 52    & \textbf{22} & \textbf{22} & \textbf{22} & \textbf{22} & \textbf{22} \\
ring  & \textbf{63} & \textbf{70} & N/A   & 1126  & N/A   & N/A \\
schmitfast & \textbf{46} & 71    & 82    & 69    & 67    & \textbf{64} \\
schmitslow & N/A   & N/A   & 127   & 96    & \textbf{93} & 108 \\
slowlatch & \textbf{58} & N/A   & 169   & \textbf{108} & 163   & 135 \\
toronto & \textbf{38} & N/A   & 277   & \textbf{258} & 277   & 273 \\
\midrule
\textbf{mos3} & \textbf{Ngspice} & \textbf{PPTA} & \textbf{CEPTA} & \textbf{MES} & \textbf{EI} & \textbf{UCB} \\
\midrule
arom  & \textbf{28} & N/A   & N/A   & N/A   & N/A   & N/A \\
b330  &  N/A      & N/A   & N/A   & N/A   & N/A   & N/A \\
counter &   N/A     & \textbf{22} & \textbf{22} & \textbf{22} & \textbf{22} & \textbf{22} \\
gm1   & 1380  & N/A   & 76    & \textbf{74} & \textbf{74} & \textbf{74} \\
gm2   & 59    & N/A   & 70    & \textbf{47} & 49    & \textbf{47} \\
gm3   & 61    & N/A   & 66    & 53    & 51    & \textbf{50} \\
gm17  & \textbf{35} & N/A   & 212   & \textbf{185} & 191   & 192 \\
gm19  & \textbf{39} & N/A   & N/A   & \textbf{256} & 1744  & 267 \\
jge   & \textbf{719} & N/A   & 1215  & 826   & \textbf{778} & 801 \\
mike2 & 1208  & N/A   & 189   & 78    & 80    & \textbf{57} \\
rich3 & \textbf{28} & N/A   & N/A   & N/A   & N/A   & N/A \\
todd3 & \textbf{55} & N/A   & 554   & 219   & \textbf{105} & 133 \\
\bottomrule
\end{tabular}%
\end{adjustbox}
\end{table}

\subsection{Optimal Predictions For Unseen Simulations}
\begin{figure*}[t]
    \centering
    \vspace{-0.1in}
    \includegraphics[scale=0.3]{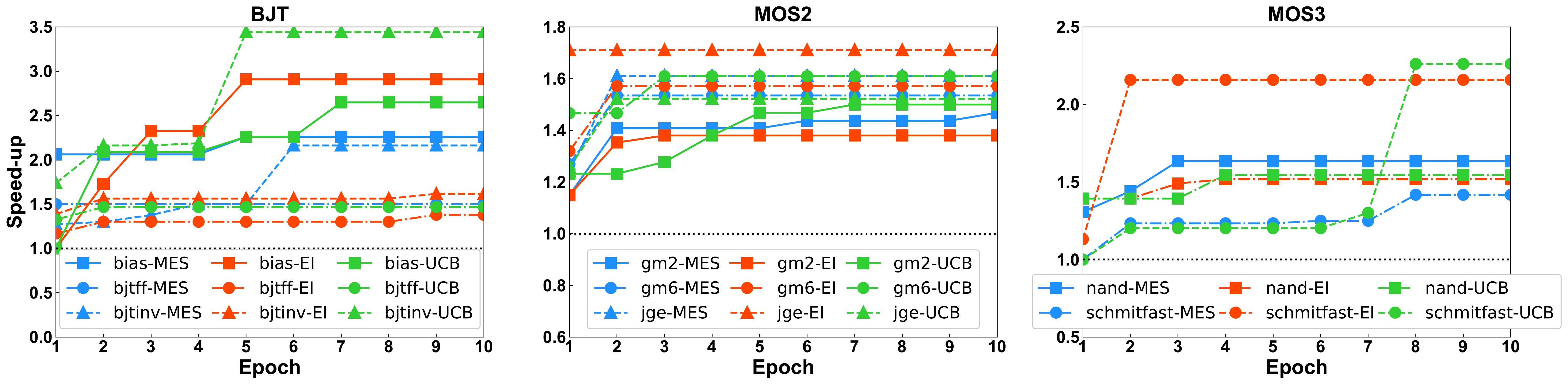}
    \caption{Speed-up for unseen BJT-, MOS2-, and MOS3-type circuit simulations.}
    \label{fig_unseen}
\end{figure*}

In this experiment, we pick the circuit with large potential for improvements among all types of circuits, i.e., $\mathcal{T}$=\{bias, bjtff, bjtinv, gm2, gm6, jge, nand, schmitfast\} and use them as testing simulations for \ours.
Specifically, simulations that are not in $\mathcal{T}$ are used as the training simulations and used as input for Algorithm \ref{algo1}. 
At the end of each Epoch, we use \ours to predict solver parameters for $\mathcal{T}$ and evaluate their speed-ups. 
We emphasis that the evaluations of $\mathcal{T}$ are never updated to \ours. They are strictly treated as testing data. 
The results are shown in Fig.~\ref{fig_unseen}.
In this case we do not compare \ours with a random search optimization because there is no way for it to predict the optimal solver parameters. This is indeed one of the main novelty of \ours.

As we can see in Fig.~\ref{fig_unseen}, \ours can further improve the SPICE speed-up with increasing number of epochs even the circuits have never been seen by the system.
This essentially indicates that the DNN feature extractions indeed work with \ours as expected such that knowledge from the training circuits can be transferred to the testing circuits.
The BJT-type and MOS3-type circuit simulations show a larger potential for improvements whereas the improvement for the MOS2-type circuit simulation is less significant.
The EI acquisition shows a stable improvement with only three epochs for all three types of simulations in this experiment.
The MES, on the other hand, shows a slower improvement.

\subsection{Monte-Carlo Accelerations}
\begin{figure}[]
    \centering
    \includegraphics[scale=0.5]{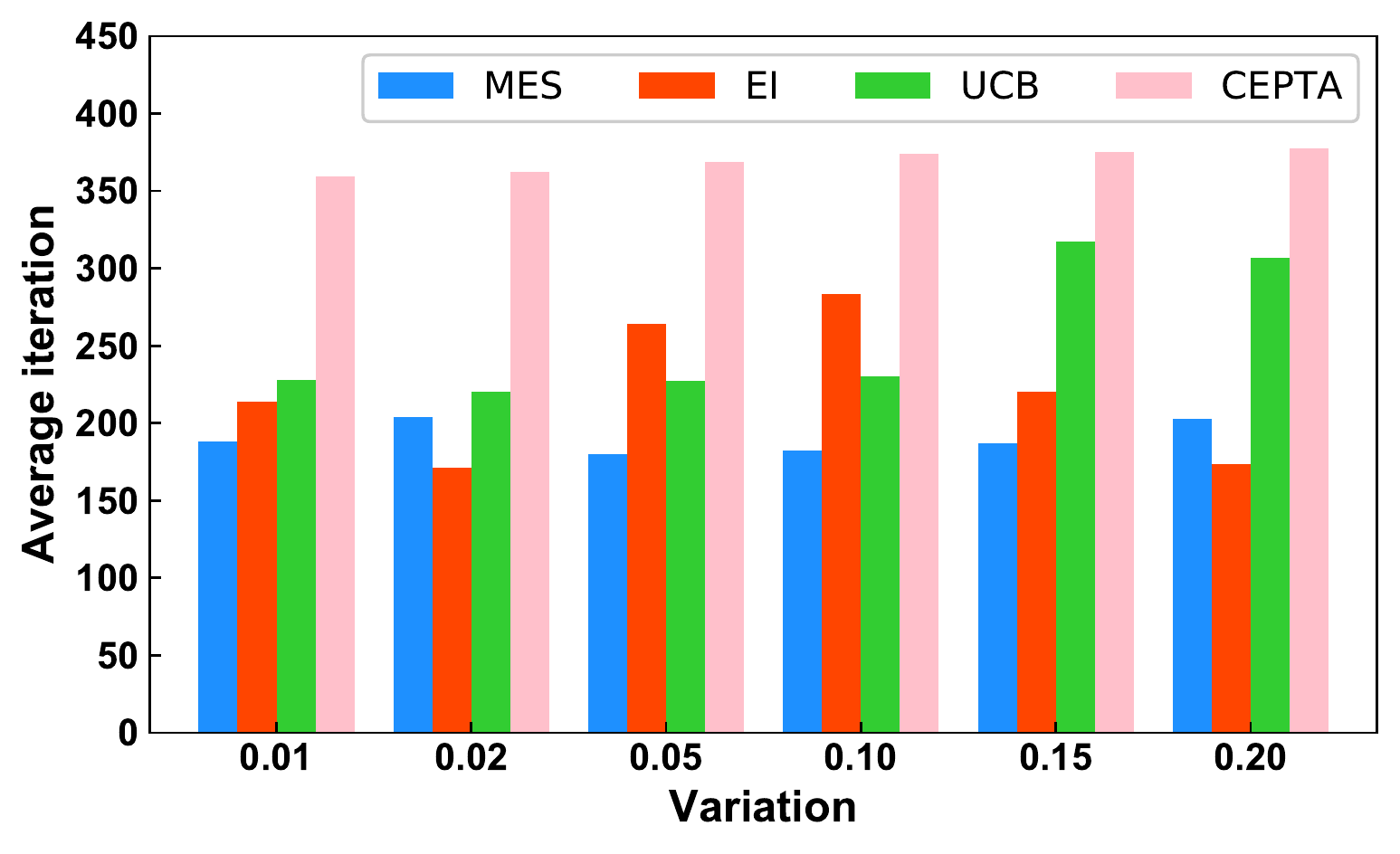}
    \caption{Average number of iterations (over bias, bjtinv, and bjtff) for Monte-Carlo simulations with different variations.}
    \label{fig_mcBar}
\end{figure}

Lastly, we assess \ours in Monte-Carlo Accelerations as described in Algorithm~\ref{algo2}. 
Similarly, we use the BJT circuit with potential for improvements, i.e., bias, bjtinv, and bjtff as testing example.
The Monte-Carlo simulation is designed to analyze the statistical properties when all registers in a circuit have independent $\{1\%, 2\%, 5\%, 10\%, 20\%\}$ variation of normal distribution, i.e., $R=R_{original}*\mathcal{N}(1,variation^2)$.
For each variation set, each method is tested on the same 1000 random sampled netlists to provide a fair comparison.
Since \ours is error-free as discussed, we did not show the statistical results but focus on the run-time statistics. 
We firstly show the number of non-convergence (\#NC), the mean, and the standard deviation (STD) iterations for 6000 total simulations in Table \ref{tab_mcIter}.
We can see clearly that \ours always converges and always provides minimal iterations. Among different acquisition functions, \ours with MES consistently show the best performance, with is consistent with the observation in previous experiments. 
CEPTA always has the lowest standard deviation of iterations. We argue that what matters most is the total iterations not the deviation for the run-time. Also, \ours can overcome a few non-converge simulations in the bias circuits. The other PTA solvers are way worse than \ours and CEPTA, which is consistent with previous results.
\begin{table}[]
    \caption{Monte-Carlo simulation statistics}
    \label{tab_mcIter}
    \begin{adjustbox}{width=1\columnwidth,center}
    \begin{tabular}{c|c|ccccccc}
        \toprule
        Circuit &       & DPTA  & RPTA  & PPTA  & CEPTA & MES   & EI    & UCB \\
        \midrule
        \multirow{3}[2]{*}{bias} & \#NC  & 3913  & 88    & 5710  & 9     & \textbf{0 } & \textbf{0 } & \textbf{0 } \\
              & Mean  & 9993  & 722   & 5644  & 913   & \textbf{346 } & 359   & 484  \\
              & STD   & 29980  & 4342  & 8395  & \textbf{51 } & 191   & 465   & 2964  \\
        \midrule
        \multirow{3}[2]{*}{bjtinv} & \#NC  & \textbf{0} & \textbf{0} & \textbf{0} & \textbf{0} & \textbf{0} & \textbf{0} & \textbf{0} \\
              & Mean  & 73.9  & 73.8  & 124.5  & 55.3  & \textbf{49.5 } & 53.8  & 48.8  \\
              & STD   & 14.4  & 14.8  & 10.4  & \textbf{3.3 } & 12.3  & 13.9  & 10.3  \\
        \midrule
        \multirow{3}[2]{*}{bjtff} & \#NC  & 6000  & 6000  & 6000  & \textbf{0} & \textbf{0} & \textbf{0} & \textbf{0} \\
              & Mean  & N/A   & N/A   & N/A   & 138   & \textbf{89} & 93    & 93 \\
              & STD   & N/A   & N/A   & N/A   & \textbf{0} & 13.5  & 20.0  & 4.8  \\
        \bottomrule
    \end{tabular}%
\end{adjustbox}
\end{table}
The average iterations number (over 6000 simulations) are shown in Fig.~\ref{fig_mcBar} without DPTA, RPTA, and PPTA due to their high non-convergence rate. 
Compared to CEPTA, we can see that whichever acquisition function improves \ours for a large margin. \ours with MES overall obtains the most stable and good performance with approximately 2x speed-up over CEPTA.

\section{Conclusion}
In this paper, a Bayesian optimization SPICE acceleration is proposed and it is demonstrated using \ours, a combination with PTA solver. By harnessing the advantages of modern machine learning techniques, \ours demonstrate a substantial improvement (up to 3.5x speed-up) over the original CEPTA solver with little extra cost.
The improvements are validated through 43 benchmark circuit simulations and Monte-Carlo simulations.

\clearpage

\bibliographystyle{IEEEtran}

\begin{thebibliography}{10}
	\providecommand{\url}[1]{#1}
	\csname url@samestyle\endcsname
	\providecommand{\newblock}{\relax}
	\providecommand{\bibinfo}[2]{#2}
	\providecommand{\BIBentrySTDinterwordspacing}{\spaceskip=0pt\relax}
	\providecommand{\BIBentryALTinterwordstretchfactor}{4}
	\providecommand{\BIBentryALTinterwordspacing}{\spaceskip=\fontdimen2\font plus
		\BIBentryALTinterwordstretchfactor\fontdimen3\font minus
		\fontdimen4\font\relax}
	\providecommand{\BIBforeignlanguage}[2]{{%
			\expandafter\ifx\csname l@#1\endcsname\relax
			\typeout{** WARNING: IEEEtran.bst: No hyphenation pattern has been}%
			\typeout{** loaded for the language `#1'. Using the pattern for}%
			\typeout{** the default language instead.}%
			\else
			\language=\csname l@#1\endcsname
			\fi
			#2}}
	\providecommand{\BIBdecl}{\relax}
	\BIBdecl
	
	\bibitem{chen2017parallel}
	X.~Chen, Y.~Wang, and H.~Yang, \emph{Parallel sparse direct solver for
		integrated circuit simulation}.\hskip 1em plus 0.5em minus 0.4em\relax
	Springer, 2017.
	
	\bibitem{akbari2018input}
	M.~Akbari, O.~Hashemipour, and F.~Moradi, ``Input offset estimation of cmos
	integrated circuits in weak inversion,'' \emph{IEEE Transactions on Very
		Large Scale Integration (VLSI) Systems}, vol.~26, no.~9, pp. 1812--1816,
	2018.
	
	\bibitem{paul2002testing}
	B.~C. Paul and K.~Roy, ``Testing cross-talk induced delay faults in static cmos
	circuit through dynamic timing analysis,'' in \emph{Proceedings.
		International Test Conference}.\hskip 1em plus 0.5em minus 0.4em\relax IEEE,
	2002, pp. 384--390.
	
	\bibitem{zhang2020efficient}
	S.~Zhang, F.~Yang, D.~Zhou, and X.~Zeng, ``An efficient asynchronous batch
	bayesian optimization approach for analog circuit synthesis,'' in \emph{2020
		57th ACM/IEEE Design Automation Conference (DAC)}.\hskip 1em plus 0.5em minus
	0.4em\relax IEEE, 2020, pp. 1--6.
	
	\bibitem{negel1975computer}
	L.~Negel, ``A computer program to simulate semiconductor circuits,''
	\emph{College of Engineering, Univ. of California Memorandum}, 1975.
	
	\bibitem{wang2009electronic}
	L.-T. Wang, Y.-W. Chang, and K.-T.~T. Cheng, \emph{Electronic design
		automation: synthesis, verification, and test}.\hskip 1em plus 0.5em minus
	0.4em\relax Morgan Kaufmann, 2009.
	
	\bibitem{huang2021machine}
	G.~{Huang}, J.~{Hu}, Y.~{He}, J.~{Liu}, M.~{Ma}, Z.~{Shen}, J.~{Wu}, Y.~{Xu},
	H.~{Zhang}, K.~{Zhong}, X.~{Ning}, Y.~{Ma}, H.~{Yang}, B.~{Yu}, H.~{Yang},
	and Y.~{Wang}, ``Machine learning for electronic design automation: A
	survey.'' \emph{arXiv: Signal Processing}, 2021.
	
	\bibitem{zhang2019efficient}
	S.~Zhang, W.~Lyu, F.~Yang, C.~Yan, D.~Zhou, X.~Zeng, and X.~Hu, ``An efficient
	multi-fidelity bayesian optimization approach for analog circuit synthesis,''
	in \emph{2019 56th ACM/IEEE Design Automation Conference (DAC)}.\hskip 1em
	plus 0.5em minus 0.4em\relax IEEE, 2019, pp. 1--6.
	
	\bibitem{liu2010an}
	B.~{Liu}, F.~V. {Fernandez}, and G.~{Gielen}, ``An accurate and efficient
	yield optimization method for analog circuits based on computing budget
	allocation and memetic search technique,'' in \emph{2010 Design, Automation
		\& Test in Europe Conference \& Exhibition (DATE 2010)}, 2010, pp.
	1106--1111.
	
	\bibitem{gunther1995dae}
	M.~G{\"u}nther and U.~Feldmann, ``The dae-index in electric circuit
	simulation,'' \emph{Mathematics and Computers in Simulation}, vol.~39, no.
	5-6, pp. 573--582, 1995.
	
	\bibitem{udave2012dc}
	D.~E.~C. Udave, J.~Ogrodzki, and G.~de~Anda, ``Dc large-scale simulation of
	nonlinear circuits on parallel processors,'' \emph{International Journal of
		Electronics and Telecommunications}, vol.~58, no.~3, pp. 285--295, 2012.
	
	\bibitem{DBLP:journals/tcad/RoychowdhuryM06}
	J.~S. Roychowdhury and R.~C. Melville, ``Delivering global {DC} convergence for
	large mixed-signal circuits via homotopy/continuation methods,'' \emph{{IEEE}
		Trans. Comput. Aided Des. Integr. Circuits Syst.}, vol.~25, no.~1, pp.
	66--78, 2006.
	
	\bibitem{ter2012robust}
	E.~J.~W. ter Maten, T.~G. Beelen, A.~de~Vries, and M.~van Beurden, ``Robust
	time-domain source stepping for dc-solution of circuit equations,'' in
	\emph{Scientific Computing in Electrical Engineering (SCEE 2012), September
		11-14, 2012, Zurich, Switzerland}, 2012, pp. 39--40.
	
	\bibitem{DBLP:journals/tcad/UshidaYNKI02}
	A.~Ushida, Y.~Yamagami, Y.~Nishio, I.~Kinouchi, and Y.~Inoue, ``An efficient
	algorithm for finding multiple {DC} solutions based on the spice-oriented
	newton homotopy method,'' \emph{{IEEE} Trans. Comput. Aided Des. Integr.
		Circuits Syst.}, vol.~21, no.~3, pp. 337--348, 2002.
	
	\bibitem{rezvani2017synthesis}
	M.~A. Rezvani, M.~A. Asli, S.~Khandan, H.~Mousavi, and Z.~S. Aghbolagh,
	``Synthesis and characterization of new nanocomposite ctab-pta@ cs as an
	efficient heterogeneous catalyst for oxidative desulphurization of
	gasoline,'' \emph{Chemical Engineering Journal}, vol. 312, pp. 243--251,
	2017.
	
	\bibitem{wu2014pta}
	X.~Wu, Z.~Jin, D.~Niu, and Y.~Inoue, ``A pta method using numerical integration
	algorithms with artificial damping for solving nonlinear dc circuits,''
	\emph{Nonlinear Theory and Its Applications, IEICE}, vol.~5, no.~4, pp.
	512--522, 2014.
	
	\bibitem{jin2015effective}
	Z.~Jin, X.~Wu, D.~Niu, X.~Guan, and Y.~Inoue, ``Effective ramping algorithm and
	restart algorithm in the spice3 implementation for dpta method,''
	\emph{Nonlinear Theory and Its Applications, IEICE}, vol.~6, no.~4, pp.
	499--511, 2015.
	
	\bibitem{DBLP:journals/ieicet/YuISHH07}
	H.~Yu, Y.~Inoue, K.~Sako, X.~Hu, and Z.~Huang, ``An effective {SPICE3}
	implementation of the compound element pseudo-transient algorithm,''
	\emph{{IEICE} Trans. Fundam. Electron. Commun. Comput. Sci.}, vol. 90-A,
	no.~10, pp. 2124--2131, 2007.
	
	\bibitem{mockus2012bayesian}
	J.~Mockus, \emph{Bayesian approach to global optimization: theory and
		applications}.\hskip 1em plus 0.5em minus 0.4em\relax Springer Science \&
	Business Media, 2012, vol.~37.
	
	\bibitem{kennedy2001bayesian}
	M.~C. Kennedy and A.~O'Hagan, ``Bayesian calibration of computer models,''
	\emph{Journal of the Royal Statistical Society: Series B (Statistical
		Methodology)}, vol.~63, no.~3, pp. 425--464, 2001.
	
	\bibitem{rasmussen2006gaussian}
	C.~E. Rasmussen and C.~K.~I. Williams, \emph{Gaussian Processes for Machine
		Learning}.\hskip 1em plus 0.5em minus 0.4em\relax MIT Press, 2006.
	
	\bibitem{jones1998efficient}
	D.~R. {Jones}, M.~{Schonlau}, and W.~J. {Welch}, ``Efficient global
	optimization of expensive black-box functions,'' \emph{Journal of Global
		Optimization}, vol.~13, no.~4, pp. 455--492, 1998.
	
	\bibitem{srinivas2010gaussian}
	N.~{Srinivas}, A.~{Krause}, M.~{Seeger}, and S.~M. {Kakade}, ``Gaussian process
	optimization in the bandit setting: No regret and experimental design,'' in
	\emph{Proceedings of the 27th International Conference on Machine Learning},
	2010, pp. 1015--1022.
	
	\bibitem{wang2017max}
	Z.~{Wang} and S.~{Jegelka}, ``Max-value entropy search for efficient bayesian
	optimization,'' in \emph{Proceedings of the 34th International Conference on
		Machine Learning - Volume 70}, 2017, pp. 3627--3635.
	
	\bibitem{scott2011the}
	W.~{Scott}, P.~{Frazier}, and W.~B. {Powell}, ``The correlated knowledge
	gradient for simulation optimization of continuous parameters using gaussian
	process regression,'' \emph{Siam Journal on Optimization}, vol.~21, no.~3,
	pp. 996--1026, 2011.
	
	\bibitem{hernandez-lobato2014predictive}
	J.~M. {Hernandez-Lobato}, M.~W. {Hoffman}, and Z.~{Ghahramani}, ``Predictive
	entropy search for efficient global optimization of black-box functions,'' in
	\emph{Advances in Neural Information Processing Systems 27}, vol.~27, 2014,
	pp. 918--926.
	
	\bibitem{lyu2018batch}
	W.~Lyu, F.~Yang, C.~Yan, D.~Zhou, and X.~Zeng, ``Batch bayesian optimization
	via multi-objective acquisition ensemble for automated analog circuit
	design,'' in \emph{International conference on machine learning}.\hskip 1em
	plus 0.5em minus 0.4em\relax PMLR, 2018, pp. 3306--3314.
	
	\bibitem{ma2019high}
	Y.~{Ma}, H.~{Ren}, B.~{Khailany}, H.~{Sikka}, L.~{Luo}, K.~{Natarajan}, and
	B.~{Yu}, ``High performance graph convolutional networks with applications in
	testability analysis,'' in \emph{Proceedings of the 56th Annual Design
		Automation Conference 2019 on}, 2019, p.~18.
	
	\bibitem{zhang2019bayesian}
	S.~{Zhang}, W.~{Lyu}, F.~{Yang}, C.~{Yan}, D.~{Zhou}, and X.~{Zeng}, ``Bayesian
	optimization approach for analog circuit synthesis using neural network,'' in
	\emph{2019 Design, Automation \& Test in Europe Conference \& Exhibition
		(DATE)}, 2019, pp. 1463--1468.
	
	\bibitem{lecun2015deep}
	Y.~{LeCun}, Y.~{Bengio}, and G.~{Hinton}, ``Deep learning,'' \emph{Nature},
	vol. 521, no. 7553, pp. 436--444, 2015.
	
	\bibitem{wilson2016deep}
	A.~G. Wilson, Z.~Hu, R.~Salakhutdinov, and E.~P. Xing, ``Deep kernel
	learning,'' in \emph{Artificial Intelligence and Statistics}, 2016, pp.
	370--378.
	
	\bibitem{bornn2012modeling}
	L.~Bornn, G.~Shaddick, and J.~V. Zidek, ``Modeling nonstationary processes
	through dimension expansion,'' \emph{Journal of the American Statistical
		Association}, vol. 107, no. 497, pp. 281--289, 2012.
	
	\bibitem{adams2008gaussian}
	R.~P. {Adams} and O.~{Stegle}, ``Gaussian process product models for
	nonparametric nonstationarity,'' in \emph{Proceedings of the 25th
		international conference on Machine learning}, 2008, pp. 1--8.
	
	\bibitem{slaets2013warping}
	L.~{Slaets}, G.~{Claeskens}, and B.~W. {Silverman}, ``Warping functional data
	in r and c via a bayesian multiresolution approach,'' \emph{Journal of
		Statistical Software}, vol.~55, no.~1, pp. 1--22, 2013.
	
	\bibitem{kingma2014auto}
	D.~P. {Kingma} and M.~{Welling}, ``Auto-encoding variational bayes,'' in
	\emph{ICLR 2014 : International Conference on Learning Representations (ICLR)
		2014}, 2014.
	
	\bibitem{xing2016manifold}
	W.~Xing, V.~Triantafyllidis, A.~Shah, P.~Nair, and N.~Zabaras, ``Manifold
	learning for the emulation of spatial fields from computational models,''
	\emph{Journal of Computational Physics}, vol. 326, pp. 666--690, 2016.
	
	\bibitem{titsias2009variational}
	M.~K. Titsias, ``Variational learning of inducing variables in sparse gaussian
	processes,'' in \emph{International Conference on Artificial Intelligence and
		Statistics}, 2009, pp. 567--574.
	
	\bibitem{barby1993circuitsim93}
	J.~{Barby} and R.~{Guindi}, ``Circuitsim93: A circuit simulator benchmarking
	methodology case study,'' in \emph{Sixth Annual IEEE International ASIC
		Conference and Exhibit}, 1993, pp. 531--535.
	
	\bibitem{780151}
	E.~Yilmaz and M.~Green, ``Some standard spice dc algorithms revisited: why does
	spice still not converge?'' in \emph{1999 IEEE International Symposium on
		Circuits and Systems (ISCAS)}, vol.~6, 1999, pp. 286--289 vol.6.
	
\end{thebibliography}

\end{document}